\documentclass[letterpaper, 10 pt, conference]{ieeeconf}  % Comment this line out if you need a4paper
\usepackage{amsmath,amsfonts}
\usepackage{algorithm}
\usepackage{array}
\usepackage[caption=false,font=normalsize,labelfont=sf,textfont=sf]{subfig}
\usepackage{textcomp}
\usepackage{stfloats}
\usepackage{url}
\usepackage{verbatim}
\usepackage{graphicx}
\usepackage{cite}
\usepackage{algorithm}
\usepackage{algpseudocode}

\usepackage{amsmath}

\IEEEoverridecommandlockouts                              % This command is only needed if 
\title{\LARGE \bf
Tacchi 2.0: A Low Computational Cost and Comprehensive Dynamic Contact Simulator for Vision-based Tactile Sensors}

\author{Yuhao Sun$^{1}$, Shixin Zhang$^{2}$, Wenzhuang Li$^{3}$, Jie Zhao$^{4}$, Jianhua Shan$^{4}$, Zirong Shen$^{5}$\\ Zixi Chen$^{6}$, Fuchun Sun$^{7}$, Di Guo$^{1}$ and Bin Fang$^{1*}$% <-this % stops a space
\thanks{This work was jointly supported by the National Natural Science Foundation of China (Grant No.62173197, No.U22B2042).}
\thanks{Yuhao Sun and Shixin Zhang contributed equally to this work. \emph{*Corresponding author: Bin Fang.}}
\thanks{$^{1}$School of Artificial Intelligence, Beijing University of Posts and Telecommunications, Beijing 100083, China. Email: yuhaosun@bupt.edu.cn, fangbin1120@bupt.edu.cn.}%
\thanks{$^{2}$School of Engineering and Technology, China University of Geosciences (Beijing), Beijing 100083, China. Email: zhangshixin@email.cugb.edu.cn.}%
\thanks{$^{3}$School of International, Beijing University of Posts and Telecommunications, Beijing 100876, China. Email: lwz1195467994@bupt.edu.cn.}
\thanks{$^{4}$School of Mechanical Engineering, Anhui University of Technology, China.}
\thanks{$^{5}$Zhili College, Tsinghua University, Beijing 100084, China. Email: shenzr21@mails.tsinghua.edu.cn.}
\thanks{$^{6}$Biorobotics Institute and the Department of Excellence in Robotics and AI, Scuola Superiore Sant’Anna, 56127 Pisa, Italy. Email: zixi.chen@santannapisa.it.}
\thanks{$^{7}$Institute for Artificial Intelligence, Department of Computer Science and Technology, Beijing National Research Center for Information Science and Technology, Tsinghua University, Beijing 100084, China. Email: fcsun@mail.tsinghua.edu.cn.}
}

\begin{document}

\maketitle
\thispagestyle{empty}
\pagestyle{empty}

%%%%%%%%%%%%%%%%%%%%%%%%%%%%%%%%%%%%%%%%%%%%%%%%%%%%%%%%%%%%%%%%%%%%%%%%%%%%%%%%
\begin{abstract}
With the development of robotics technology, some tactile sensors, such as vision-based sensors, have been applied to contact-rich robotics tasks. However, the durability of vision-based tactile sensors significantly increases the cost of tactile information acquisition. Utilizing simulation to generate tactile data has emerged as a reliable approach to address this issue. While data-driven methods for tactile data generation lack robustness, finite element methods (FEM) based approaches require significant computational costs. To address these issues, we integrated a pinhole camera model into the low computational cost vision-based tactile simulator Tacchi that used the Material Point Method (MPM) as the simulated method, completing the simulation of marker motion images. We upgraded Tacchi and introduced Tacchi 2.0. This simulator can simulate tactile images, marked motion images, and joint images under different motion states like pressing, slipping, and rotating. Experimental results demonstrate the reliability of our method and its robustness across various vision-based tactile sensors.
\end{abstract}

%%%%%%%%%%%%%%%%%%%%%%%%%%%%%%%%%%%%%%%%%%%%%%%%%%%%%%%%%%%%%%%%%%%%%%%%%%%%%%%%
\section{INTRODUCTION}
Learning-based methods have successful applications in various robotics tasks, including perception \cite{DG21}, detection \cite{SZ23}, and manipulation \cite{YS21}. As robotic technology advances, the need for highly robust robot manipulation systems increases the demand for quality datasets. However, real-world data collection is costly, time-consuming, and carries risks of wear and accidents to robots. Therefore, leveraging simulation to efficiently and cost-effectively generate the data for training agents has emerged as a dependable strategy.

\begin{figure}[ht]
\centering
\includegraphics[width=3.4in]{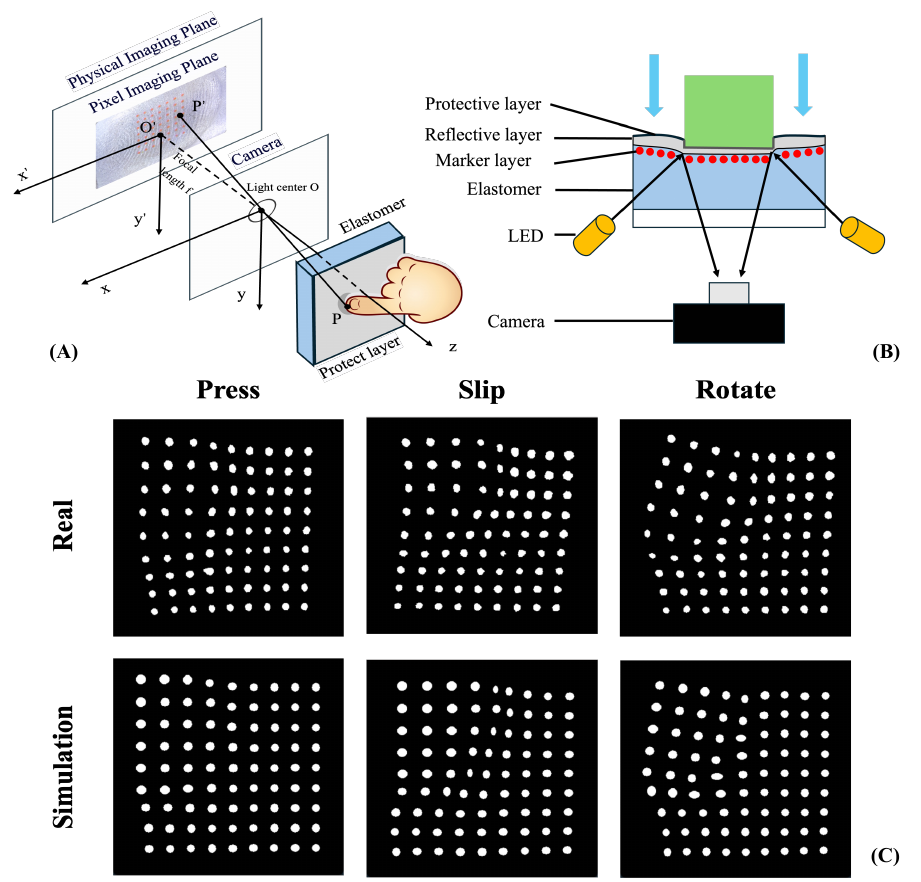}
\caption{The marker points move with the elastomer when pressed, and these displacements are captured by a camera and presented as images. This paper uses the MPM and a camera model to simulate the elastomer deformation and the camera's capture process, respectively. (\textbf{A}): Pinhole camera model schematic diagram. (\textbf{B}): Schematic diagram of vision-based tactile sensor. (\textbf{C}): Marker motion simulation effects of pressing, sliding, and rotating.}
\label{fig1}
\end{figure}

Tactile sensation plays a crucial role in robot contact control, and in recent years, many tactile sensors have been developed \cite{YS21}. Vision-based tactile sensors utilize cameras to capture deformations in elastomers, generating high-resolution tactile images. Zhang et al. \cite{SZ22} classify vision-based tactile sensors into two types based on whether or not a marker layer is used. Among them, the sensor with a marker layer is more suitable for tasks such as force measurement, while the sensor without a marker layer is more suitable for tasks such as texture detection. Some researchers have applied vision-based tactile sensors such as GelSight \cite{WY17}, TactTip \cite{BW18}, and \cite{BF18} to robotic manipulation tasks \cite{AC20, ZZ22}. To reduce the cost of dataset collection, people have begun to focus on the simulation of vision-based tactile sensors, as mentioned in \cite{YW21, SC19, SZ24}. However, the complex physical deformation and light models of vision-based tactile sensors increase the difficulty of simulation \cite{YW21}. This poses a significant challenge to the reliability and efficiency of the simulation.

Currently, simulation for vision-based tactile sensors primarily involves tactile image simulation and marker motion simulation corresponding to whether there is a marker layer. Data-driven methods have been applied to both simulations\cite{CR22, ZY24}, but this approach lacks robustness and involves high data collection costs. In contrast, physics-based simulation methods such as Finite Element Method (FEM)\cite{si2022taxim} and Material Point Method (MPM)\cite{ZC23} offer higher robustness with lower data collection costs. However, FEM-based simulations for tactile and marker motion images require significant computational resources. MPM-based simulations are currently limited to tactile image simulation.

To address these issues, building upon our previous work Tacchi\cite{ZC23}, we introduced a universal simulation method for marker motion images utilizing the MPM and camera models. Simultaneously, we integrated this simulation method into Tacchi, thereby completing an upgrade from Tacchi to Tacchi 2.0, which includes tactile image, marker motion image, and preliminary joint image simulation. Our main contributions are as follows:

\textbf{1)} We proposed an efficient marker motion image simulation method using MPM and camera models. This method can be applied to different vision-based tactile sensors by calibrating the cameras used by different sensors.

\textbf{2)} We completed the simulation and evaluation of marker motion images under pressing, slipping, and rotating. Building upon Tacchi, we completed and introduced Tacchi 2.0, a simulator capable of the tactile image, marker motion image, and joint image simulation under different motion states such as pressing, slipping, and rotating.

The rest of the paper is structured as follows: Section \ref{sec:B} introduces related work on vision-based tactile simulation; Section \ref{sec:C} introduces the upgrades and implementation methods of Tacchi 2.0; Section \ref{sec:D} presents the setup of experiments and results; Section \ref{sec:E} summarizes this work and provides outlooks for future work.

%%%%%%%%%%%%%%%%%%%%%%%%%%%%%%%%%%%%%%%%%%%%%%%%%%%%%%%%%%%%%%%%%%%%%%%%%%%%%%%%%%
\section{RELATED WORK}
\label{sec:B}
\subsection{Elastomer Simulation}
The key to physics-based vision-based tactile sensor simulation lies in elastomer simulation. Most elastomers are made of silicone, which exhibits elastic deformation characteristics, thereby increasing the complexity and cost of simulation.

To reduce the complexity of simulating elastic deformations, Gomes\cite{DG21} utilized a method involving Gaussian Kernel to smooth depth information, thereby reconstructing the mapping of the elastic objects. Similarly, in a related manner,\cite{YW21} utilized object depth in contact regions and employed kernel smoothing methods in non-contact areas. Additionally, Pybullet has been employed in the simulation of elastic object deformations\cite{WS22}. While these methods simulate tactile images, they only generate depth maps through smoothing techniques, lacking physical significance.

To achieve realistic simulations, researchers are increasingly focusing on physics-based methods such as FEM and MPM. Sferrazza\cite{SC19} analyzed contact forces using FEM. However, FEM requires significant computational resources, which can decrease simulation efficiency. Simultaneously, in the FEM, each element is connected with the other ones, and deformation simulation may also lead to mesh distortion problems like irregular meshes or even negative element volume\cite{LEE93}. Flexible deformation simulation based on MPM has garnered attention\cite{SZ24, ZC23}. This method utilizes particles to represent objects and employs a virtual grid for contact simulation, exchanging object information between the grid and particles during the simulation process. Compared to other physics-based methods like FEM, MPM maintains grid integrity due to its independent components. Therefore, we continue the previous work by utilizing MPM as the simulation method for elastomers.

\subsection{Marker Motion Simulation}
Using tactile images directly as observation to achieve transfer between the simulation and the real world is challenging\cite{SYH24}. The propagation of light in vision-based tactile sensors is a complex process, involving the transmission of light through acrylic and silicone, leading to reflection and refraction. Additionally, the influence of external vision-based factors further complicates the simulation. Therefore, using tactile images directly as observation poses challenges such as low data generation efficiency and significant gaps between the simulation and the real world.

The marker layer is an array of points embedded within a sensor, implicitly containing contact information \cite{WY17}. Using a dataset with marker motion information to train a generative model is feasible. Kim et al. \cite{kim2023marker} used GAN to generate the marker motion image reliability and efficiency. Ou et al. \cite{ou2024marker} used the diffusion model to change the model between tactile and marker motion images. However, learning-based methods need a large amount of tactile data and are difficult to be robust to different sensors.

Unlike learning-based marker motion simulation, the physics-based method can simulate different vision-based tactile sensors by adjusting parameters. Si et al. \cite{si2022taxim} used the FEM, linear displacement relationship, and superposition principle to simulate marker motion images. However, the efficiency of FEM is low and not suitable for efficient data generation. Zhao et al. \cite{ZY24} developed an efficient marker motion generation method called FOTS, which simulates only the motion of the marker, thus missing out on the information contained in the motion images of the marker.

To overcome these issues, we employed MPM combined with a pinhole camera model to simulate the motion image of the marker. The entire process is based on physics-based methods, showcasing strong robustness. Furthermore, we continued to utilize the efficient and parallel programming language Taichi, which is known for its lower computational costs.

%%%%%%%%%%%%%%%%%%%%%%%%%%%%%%%%%%%%%%%%%%%%%%%%%%%%%%%%%%%%%%%%%%%%%%%%%%%%%%%%%%

\section{METHOD}
\label{sec:C}
\subsection{Tacchi}
In our previous work, we developed the vision-based tactile sensor simulator named Tacchi\cite{ZC23}. In this work, we continued to utilize particles to represent objects, where the movement of particles can simulate the deformation of elastomer proposed by MPM. Each particle contains contact information such as mass, velocity, and deformation. Additionally, within the simulation environment, there is a fixed grid where particles exchange contact information with nearby grid nodes. The particles move at each step based on the exchanged particle velocities. Thanks to the use of particles and grids, MPM can utilize both particle-based and grid-based simulation methods simultaneously. It includes five steps: initialization, particle-to-grid, grid operations, grid-to-particle, and particle operations. These components encompass the entire process of contact information exchange, deformation simulation, and motion. For specific details, please refer to\cite{ZC23}. We refer to the coordinates in which the simulator is located as the world coordinates. Therefore, in each step of the simulation, we can obtain the coordinates representing the elastomer's particles in the world coordinate and their corresponding changes.

Tacchi 2.0 builds upon the foundation of Tacchi by incorporating a pinhole camera model. This enhancement allows Tacchi 2.0 to utilize reliable geometric relationships to transform three-dimensional information into two-dimensional images, facilitating the simulation of marker motion images. Consequently, Tacchi 2.0 can provide tactile images, marker motion images, and joint image simulation. Additionally, Tacchi 2.0 inherits the strengths of Tacchi, such as low computational resource consumption and strong compatibility with robot simulation platforms.

\begin{figure*}[t]
\centering
\includegraphics[width=6.4in]{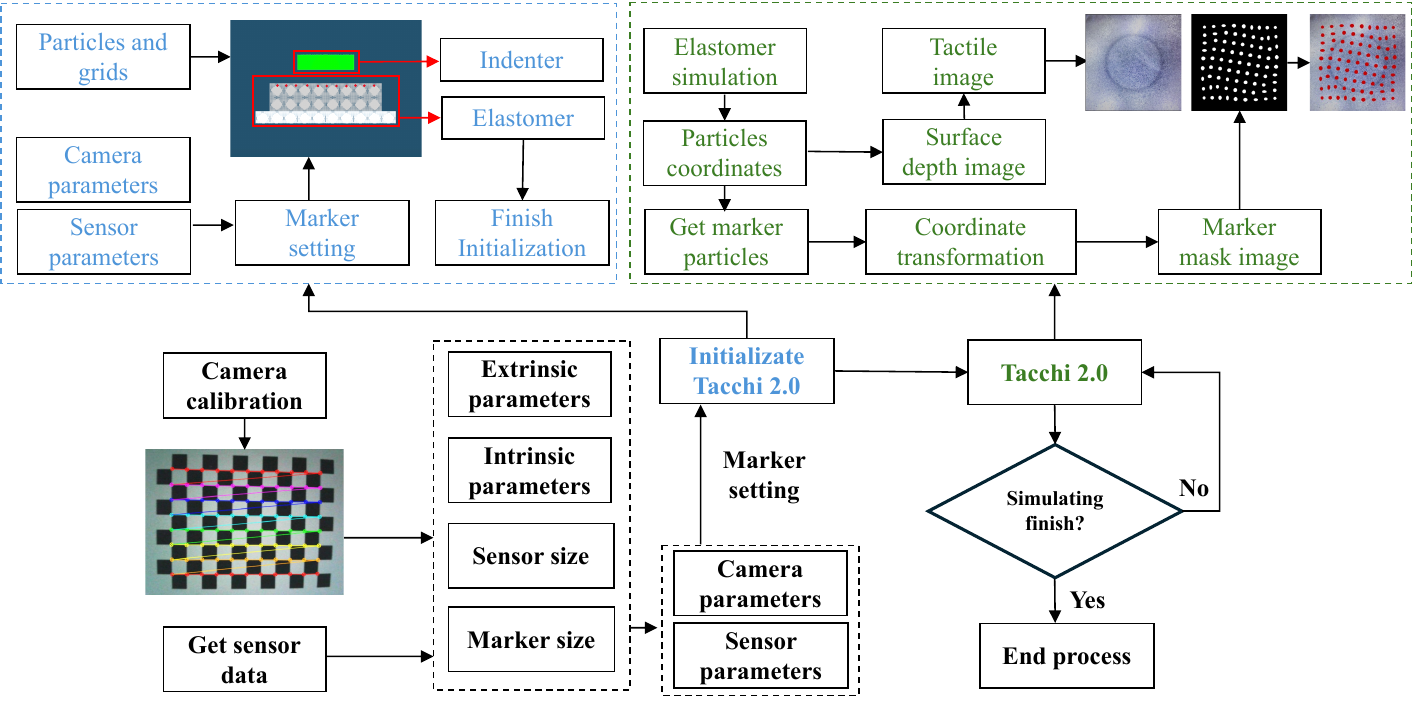}
\caption{Tacchi 2.0 overall flow chart. The overall flow chart of Tacchi 2.0, including the camera calibration module, Tacchi 2.0 initialization module, and Tacchi 2.0 calculate module.}
\label{fig2}
\end{figure*}

\subsection{Establishment of Tacchi 2.0} 
Vision-based tactile sensors utilize cameras to capture the deformation states of elastomer\cite{WY17}. In this process, the camera module maps coordinate points of the elastomer in the three-dimensional world to the two-dimensional image plane, ultimately generating the corresponding image. This process can be described using a geometric model\cite{ZZ00}, with the most commonly used being the pinhole camera model. The pinhole model describes the geometric relationship of a beam of light passing through a pinhole and being projected onto the back of the pinhole to form an image. In the previous text, we introduced the principles of Tacchi and discussed the origin of the world coordinate representing elastomer particles. After obtaining the world coordinates of the elastomer, we can use the geometric transformation process of the pinhole camera model to convert the world coordinates to the camera image coordinates, ultimately generating the marker image. The process mainly involves camera calibration, marker setup, elastomer simulation, conversion from the world coordinate to the camera coordinate, and conversion from the camera coordinate to the image coordinate, as shown in Figure \ref{fig2}.

\begin{figure}[ht]
\centering
\includegraphics[width=2.8in]{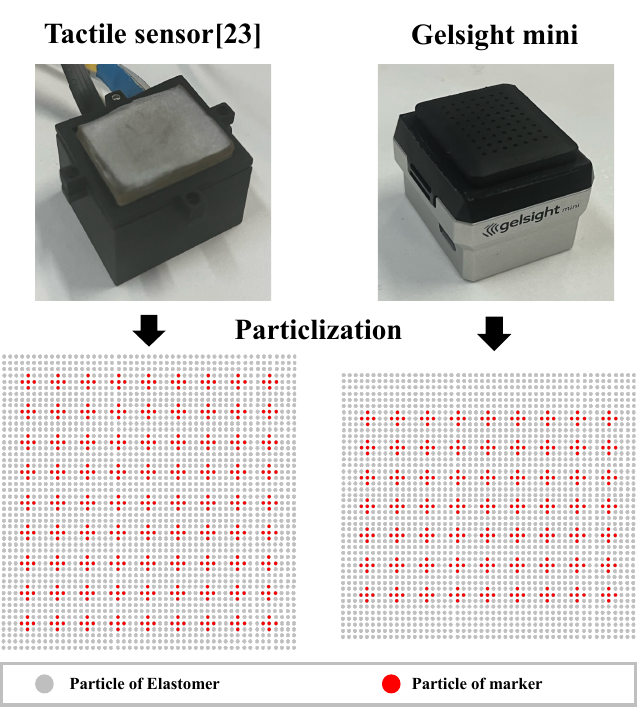}
\caption{Schematic diagram of particle formation of elastomer and representative marker point particles for different sensors.}
\label{fig3}
\end{figure}

{\bf{Camera calibration:}} Camera calibration aims to determine the external and internal parameters of the camera, which include the rotation matrix and translation vector from the world coordinate to the camera coordinate, as well as the transformation matrix from the camera coordinate to the image coordinate. Due to variations among cameras, these parameters differ. Therefore, before simulation, it is essential to calibrate the parameters of the camera being used with the help of calibration patterns and calibration algorithms. One can utilize calibration methods proposed in\cite{ZZ00}, with specific implementations already encapsulated in OpenCV or Matlab for direct use.

{\bf{Marker setting:}} Due to variations in the number of markers and their positions on the elastomer across different vision-based tactile sensors. It is necessary to mark particles according to the actual number and positions of the markers to achieve a more realistic simulation. During the simulation, the information about the marker is contained within these marked particles.

Since MPM involves discretizing objects into particles, it provides the flexibility to set up particles representing markers, as shown in Figure \ref{fig3}. We set up the markers according to the sensors designed by \cite{ZSX24} and Gelsight Mini. Multiple particles are used to represent a single marker point, enabling the depiction of marker deformations in the image.

Considering that the deformation of the marker reflects the motion state to a certain extent, the marker motion mask image has more information richness compared with the marker motion simulation that only considers the displacement \cite{si2022taxim}. In the paper, we marked a group of particles to represent a marker point. Then, we used Opencv's ellipse fitting method to fit each group of particles to achieve marker point deformation simulation.

{\bf{Elastomer simulation:}} In this work, we continue to utilize MPM and Taichi, as employed in Tacchi. The elastomer and objects are particleized. It is important to note that our simulation focuses on the deformation of the Elastomer. Hence, the rest of the sensor is disregarded. Simulating the remaining parts would be meaningless and would consume a significant amount of computational resources. Throughout the simulation process, we can obtain real-time updates of the world coordinates of the marked particles as follows:

The deformation gradient $F_p\in R^{3\times3}$ can be derived as

\begin{equation}
\label{eq1}
F_p = \frac{\partial \phi_p}{\partial x_p}(x_p),
\end{equation}
where $F_p$ is set as a three-dimensional identity matrix $I_{3 \times 3}$ in the initialization. Specifically, the mass and momentum of the nearby particles are collected. The mass $M_i$ of the $i$-th grid node is
\begin{equation}
\label{eq2}
M_i = \sum_{j \in \mathbb{G}_i} \sum_{p \in \mathbb{P}_j} {w_{jp}} m_p,
\end{equation}

The grid momentum ${MG}_i$ of the $i$-th grid node can be obtained by calculating the momentum resulted from the particle motion ${MM}_i$ and the momentum resulted from the elasticity ${ME}_i$:
\begin{equation}
\label{eq3}
{MG}_i = {MM}_i + {ME}_i.
\end{equation}

\noindent Here $MM_i$ is calculated by collecting the velocity and affine velocity of the nearby particles:
\begin{equation}
\label{eq4}
MM_i = \sum_{j \in \mathbb{G}_i} \sum_{p \in \mathbb{P}_j} w_{jp} (m_p v_p + C_p(X_j - x_p)),
\end{equation}
where $X_j$ denotes the position of the $j$-th neighboring node of the $i$-th grid node. $ ME_i $ can be obtained as in \cite{YW21} by: 
\begin{equation}
\label{eq5}
ME_i = - \triangle t \sum_{j \in \mathbb{G}_i} \sum_{p \in \mathbb{P}_j} \frac{4}{\triangle X^2} w_{jp} V_p^0 S_p(X_j - x_p),
\end{equation}
where $\triangle t$ is the time interval between two adjacent steps, $\triangle X$ is the grid node interval, $V_p^0$ denotes the initial particle volume, $S_p$ is the elasticity force for the $p$-th particle \cite{YW21}.

After obtaining $ M_i $ and $ MG_i $ using Eq. \ref{eq2} and Eq. \ref{eq3} respectively, the object velocity close to the $i$-th grid node $V_i$ can be computed as:  
\begin{equation}
\label{eq6}
V_i =\frac{MG_i}{M_i}.
\end{equation}

Suppose that the states of the $k$-th step, i.e., the velocity $v_p^{(k)}$, the affine velocity $C_p^{(k)}$, and the deformation gradient $F_p^{(k)}$, are known, $v_p^{(k+1)}$, $C_p^{(k+1)}$ and $F_p^{(k+1)}$ in the $k+1$ step can be obtained as
\begin{equation}
\label{eq7}
v_p^{(k+1)} = \sum_{i \in \mathbb{G'}_p} w_{ip}V_i^{(k)},
\end{equation}
\begin{equation}
\label{eq8}
C_p^{(k+1)} = \frac{4}{\triangle X^2} \sum_{i \in \mathbb{G'}_p} w_{ip} v_p^{(k+1)}(X_i - x_p^{(k)}),
\end{equation}
\begin{equation}
\label{eq9}
F_p^{(k+1)} = (I + \triangle t C_p^{(k+1)}) F_p^{(k)},
\end{equation}

Finally, the particles move with the above computed velocities and the position of each particle $x_p^{(k+1)}$ in the $k+1$ step is
\begin{equation}
\label{eq10}
x_p^{(k+1)} = x_p^{(k)} + \triangle t v_p^{(k+1)}.
\end{equation}

In one calculation step, MPM obtains the three-dimensional coordinates of each particle in the world coordinate where $x_{p}^{(k)}=[X_w, Y_w, Z_w]$. Then, we extract the 3D information of particles marked as markers and input them into the camera model. For a detailed calculation process of MPM, refer to\cite{ZC23}.

{\bf{Conversion from world coordinate to camera coordinate:}} By performing the camera calibration process, the external parameters of the camera can be acquired, enabling the conversion of particle positions from the world coordinate to the camera coordinate.
\begin{figure}[ht]
\centering
\includegraphics[width=3.2in]{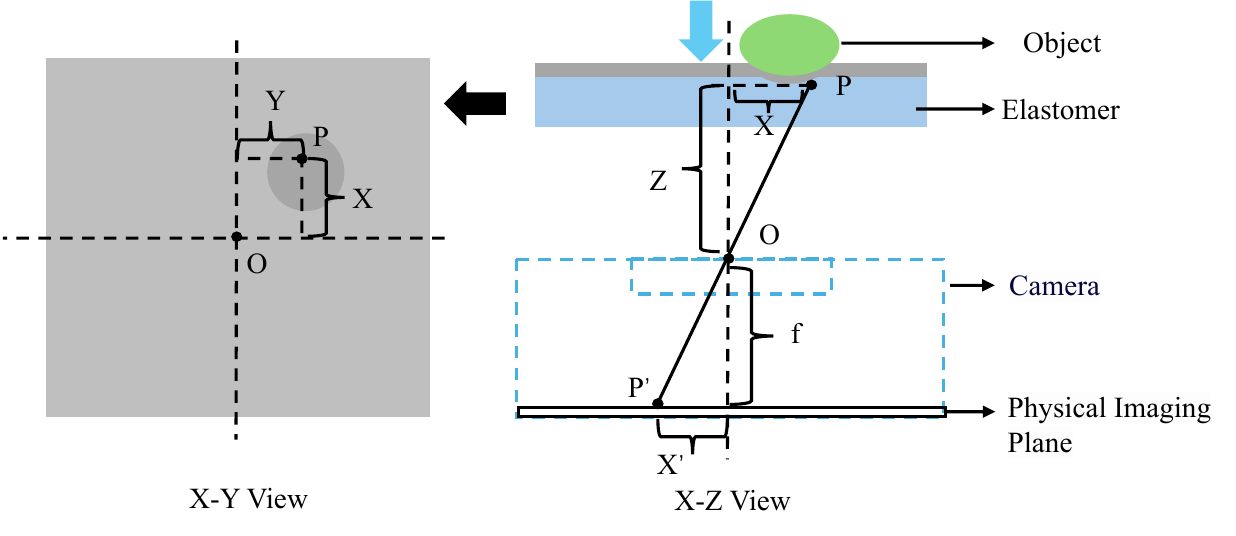}
\caption{Schematic diagram of the pinhole camera model.}
\label{fig4}
\end{figure}

\begin{equation}
\label{eq11}
x_{c}^{(k)} = Rx_p^{(k)} + t,
\end{equation}
where $R$ is the rotation matrix, $t$ is the translation vector, $x_{c}^{(k)}$ is the coordinate of a particle in the camera coordinate, and $x_p^{(k)}$ is the coordinate of a particle in the world coordinate system obtained by the simulator in equation \ref{eq8}.

{\bf{Conversion from camera coordinate to image coordinate:}} In the above process, the coordinates of particles in the camera coordinate can be obtained. Assuming $x_{c}^{(k)}=[X_c, Y_c, Z_c]$, as shown in Figure \ref{fig4}, by the principle of similar triangles, the conversion process can be obtained as:

\begin{equation}
\label{eq12}
\begin{cases}
X^\prime = f\frac{X}{Z} \\
Y^\prime = f\frac{Y}{Z}
\end{cases}
\end{equation}
In the pixel coordinate, the origin is usually the upper left corner of the image, and there is a scaling factor difference between the pixel plane and the imaging plane. The position of particles in the image coordinate is:

\begin{equation}
\label{eq13}
\begin{cases}
u = f_x\frac{X}{Z} + c_x\\
v = f_y\frac{Y}{Z} + c_y
\end{cases}
\end{equation}
where $u$ and $v$ are the positions of the corresponding points in the pixel coordinate. $f_x=\alpha f$ and $f_y=\beta f$ are the focal lengths in the directions of $x$ and $y$, respectively, where $\alpha$, and $\beta$ are the scaling ratios from the pixel plane to the imaging plane, and $f$ is the focal length of the camera. From Eq. \ref{eq10} and Eq. \ref{eq11}, the transformation relationship from the camera coordinate to the image coordinate is:

% \begin{equation}
% \label{eq14}
% \begin{bmatrix} u \\ 
% v \\
% 1 \\ 
% \end{bmatrix} 
% = \frac{1}{Z} \begin{bmatrix}
%     f_x & 0 & c_x \\
%     0   & f_y & c_y \\
%     0 & 0 & 1\\
% \end{bmatrix}\begin{bmatrix}
%     X \\
%     Y \\
%     Z \\
% \end{bmatrix}
% \end{equation}

\begin{equation}
\label{eq14}
x_I^{(k)} = \frac{K}{Z}x_c^{(k)}
\end{equation}
where $K=\begin{bmatrix}f_x & 0 & c_x \\0   & f_y & c_y \\0 & 0 & 1\\  \end{bmatrix}$ is the camera intrinsic parameter matrix, $x_I^{(k)} = \begin{bmatrix} u \\ v \\1 \\ \end{bmatrix}$, and $x_c^{(k)} = \begin{bmatrix} X_c \\ Y_c \\ Z_c \\ \end{bmatrix}$. Both the camera intrinsic parameters and the camera extrinsic parameters can be calibrated by the method proposed in \cite{ZZ00}.

After obtaining the coordinates of the marker in the pixel coordinate, a marker mask image can be obtained through segmentation and interpolation. It is important to note that the method proposed in this paper applies to all planar vision-based tactile sensors. Users only need to calibrate the camera used by the sensor beforehand and then use Tacchi 2.0 for elastomer deformation simulation to obtain simulated results of marker motion images.

\subsection{Joint Image Simulation}
In Tacchi\cite{ZC23}, our simulator was already capable of simulating tactile images. In \cite{SZ24}, we completed IMPM with the ability to simulate slipping and rotating. With the fusion of MPM and the pinhole camera model proposed in this paper, Tacchi 2.0 can simulate marker motion images under different motions such as pressing, slipping, and rotating. It is important to note that the simulation of tactile images and marker motion images both derive particle coordinates from MPM,  providing a natural fusion of adaptability and spatial alignment.

By assigning colors to markers based on real images and then adding them, we completed the initial joint simulation of tactile images and marker motion images. This enables Tacchi 2.0 to possess the capability for tactile image, marker motion image, and joint image simulation, encompassing motion forms such as pressing, slipping, and rotating. Tacchi 2.0 holds significant potential in data generation for tactile large-scale models\cite{YF24}, multimodal fusion datasets \cite{Mee22}, and large-scale robotic manipulation models.

\begin{figure*}[t]
\centering
\includegraphics[width=6.8in]{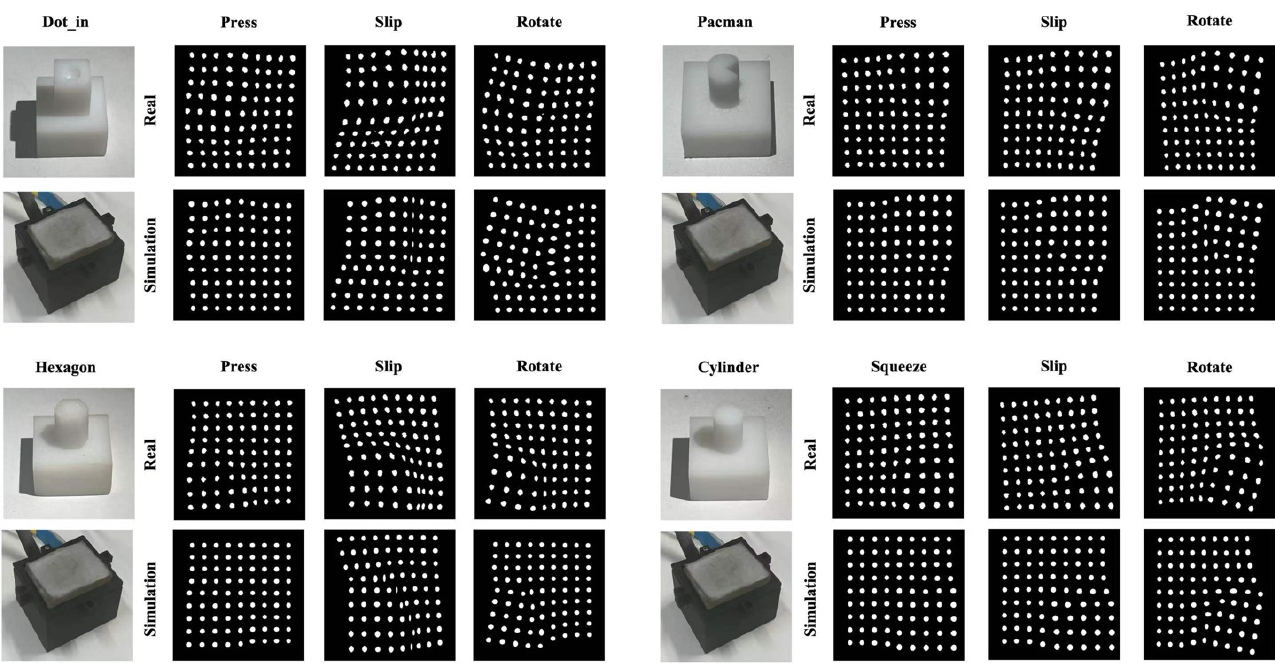}
\caption{The mask image of marker motion images under pressing, sliding, and rotating. The sensor that we used to simulate and collect real data is designed based on \cite{ZSX24}. The Size of the marker array is $9\times 9$.}
\label{fig6}
\end{figure*}

%%%%%%%%%%%%%%%%%%%%%%%%%%%%%%%%%%%%%%%%%%%%%%%%%%%%%%%%%%%%%%%%%%%%%%%%%%%%%%%%%%
\section{Experimental Design and Results}
\label{sec:D}
\subsection{Experimental Design}
In this work, we conducted experimental validation using the Gelsight Mini fabricated with spray coating techniques and the vision-based tactile sensor fabricated with gold leafing techniques\cite{SZ23}. Different indenters were employed to contact the sensors to verify the disparities between simulation and reality. 

\begin{figure}[ht]
\centering
\includegraphics[width=2.8in]{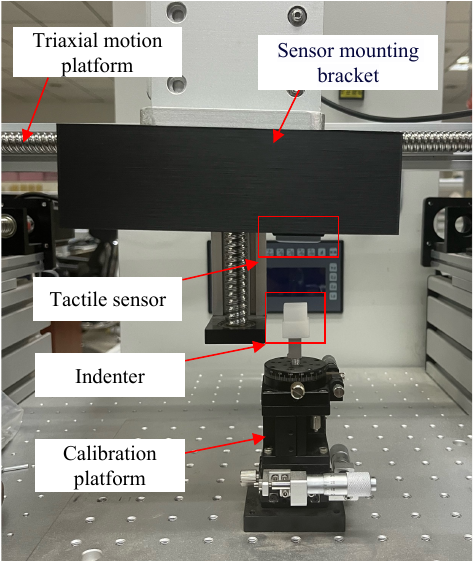}
\caption{Schematic diagram of the experimental platform and the introduction of each component.}
\label{fig5}
\end{figure}

In the real-world setup, we constructed a calibration platform as shown in Figure \ref{fig5} to carry out the respective tasks. The same indenters as in the simulation were obtained through 3D printing. In the simulation, we utilized Open3D to convert different indenters files in STL to point cloud and integrated them into the simulation environment. For the Gelsight Mini, an array of $9\times 7$ markers was used, while the array of markers for the vision-based tactile sensor designed by\cite{ZSX24} was $9\times 9$.

\subsection{Marker Motion Image Simulation}
We employed Dot\_in, Pacman, Hexagon, and Cylinder indenters to generate and collect simulation and real-world data with the same displacement. The total number of real-world data collected was $240 (20\times 4\times 3)$, with images of resolution $228\times 228$. Some of the simulation and real-world data are shown in Figure \ref{fig6}. For the algorithm of obtaining the marker motion mask image, refer to \cite{zhang2024compact}.

All of the simulation results are implemented with Taichi 1.4.0 and Python 3.8. The hardware uses an Intel Core i7-8750H processor, two 8 GB memory chips (DDR4), and one GPU (GeForce RTX 3080Ti 12 G). After initialization, the efficiency of the overall simulation can reach 37 FPS under about 48k of particles and 2M of grids.

After data collection, we used the Root Mean Squared Error (RMSE) ermse and the magnitude error emag to qualify the discrepancy between the ground-truth and predicted marker motions following \cite{kim2023marker}. The units for the relevant data are in pixels, as shown in Table \ref{table1} and Table \ref{table2}.

Based on the results from Table \ref{table1} and Table \ref{table2}, pressing achieved the best performance in both positional accuracy and shape precision. The marker in pressing motion undergoes minimal variations, thereby reducing the introduction of errors. Compared with pressing, sliding and rotating have more complex motion characteristics, which increases the error. Slide and rotate are based on pressing, which means that the corresponding dynamic movement is performed after pressing. This will increase the accumulation of errors. 

After analysis and comparison of experimental results, our method is reliable in marking dynamic contact simulation.

% MPM uses particles and grids to exchange information, but there is no information exchange between grids, which reduces the amount of information propagation. As shown in Figure \ref{fig6}, the deformation in the contact region is reliable in both simulated and real images during slipping. In the non-contact regions, there are some errors due to the issues with information transfer. This is an inherent limitation of MPM. Since the simulation of the contact region is reliable and contains the primary deformation information, this does not impact the usability of the data.

\begin{table}[!h]
\caption{The $e_{rmse}$ between the ground-truth and predicted marker motions. (PIXEL)}
\centering
\begin{tabular}{c|c c c c}
% \hline
& Mean & Press & Slip & Rotate\\
\hline
Taxim\cite{si2022taxim}&
$3.203$&
$-$&
$-$&
$-$\\
GAN\cite{kim2023marker}&
$2.297$&
$-$&
$-$&
$-$\\
Our method&
$1.313$&
$0.399$&
$2.078$&
$1.462$\\
\end{tabular}
\label{table1}
\end{table}

\begin{table}[!h]
\caption{The $e_{mag}$ between the ground-truth and predicted marker motions. (PIXEL)}
\centering
\begin{tabular}{c|c c c c}
% \hline
& Mean & Press & Slip & Rotate\\
\hline
Taxim\cite{si2022taxim}&
$3.203$&
$-$&
$-$&
$-$\\
GAN\cite{kim2023marker}&
$2.406$&
$-$&
$-$&
$-$\\
Our method&
$2.033$&
$1.275$&
$2.244$&
$2.579$\\
\end{tabular}
\label{table2}
\end{table}

In the process of marker setting, we use a group of particles to represent a marker point. We used the ellipse fitting method in OpenCV to fit each group of particles so that the deformation of the marker point can be simulated. We used the Root Mean Squared Error (RMSE) ermse and the magnitude error emag of the differences in the sizes of the minimum bounding rectangles of the marker to qualify the discrepancy between the ground-truth and predicted marker deformation. 

\begin{table}[!h]
\caption{The Shape Errors between the ground-truth and predicted marker motions. (PIXEL)}
\centering
\begin{tabular}{c|c c c}
% \hline
& Press & Slip & Rotate\\
\hline
$e_{rmse}$&
$0.296$&
$0.601$&
$0.391$\\
$e_{mag}$&
$0.321$&
$0.955$&
$0.496$\\
\end{tabular}
\label{table3}
\end{table}

Based on the results from Table \ref{table3} and Figure \ref{fig6}, our method can simulate the deformation of the marker points. The marker motion mask image has more information richness compared with the marker motion simulation that only considers the displacement \cite{si2022taxim}.

\subsection{The Robustness of Different Vision-based Tactile Sensor}

The experiments and data analysis demonstrate the reliability of our method. To showcase the robustness of different sensors, we conducted experiments using the Gelsight Mini. For detailed information about Gelsight Mini and related algorithms, please refer to GitHub: https://github.com/gelsightinc/gsrobotics. 

\begin{figure}[ht]
\centering
\includegraphics[width=3.2in]{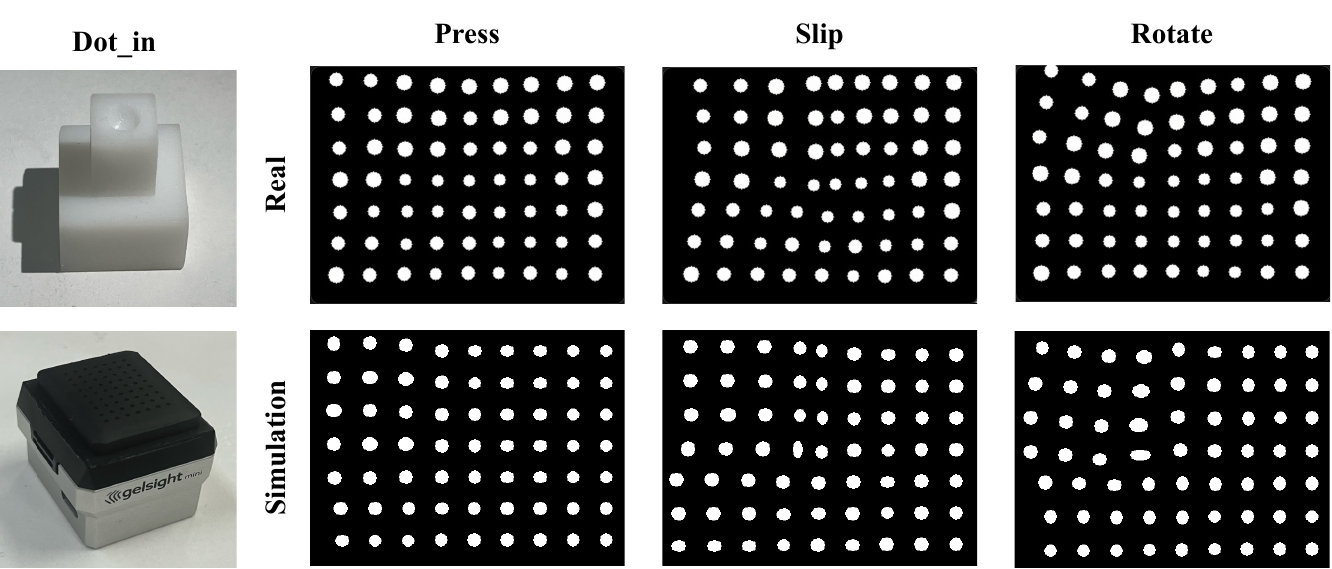}
\caption{The mask image of marker motion images under pressing, sliding, and rotating. The sensor that we used to simulate and collect real data is Gelsight Mini. The Size of the makrer array is $9\times 7$.}
\label{fig7}
\end{figure}

We utilized the dot\_in indenter to generate and collect simulation and real-world data. The results are shown in Figure \ref{fig7}.

The experimental results indicate that the marker motion image simulation method of Tacchi 2.0 for pressing, slipping, and rotating is reliable. The results from the Gelsight Mini experiments validate the robustness of Tacchi 2.0 across different cameras and sensors.

\subsection{The Joint Simulation of Tactile and Marker Motion Images}
\begin{figure}[ht]
\centering
\includegraphics[width=3.2in]{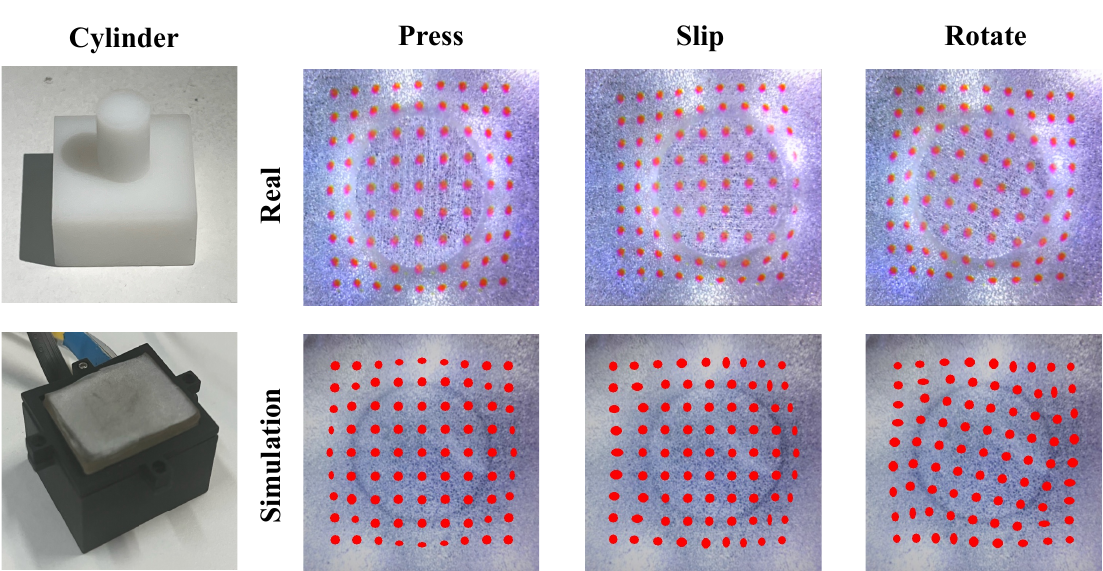}
\caption{Joint simulation results based on the sensor designed by \cite{ZSX24}. The motion simulation of the marker images is the content of this paper, and the tactile stimulation comes from IMPM\cite{SZ24}.}
\label{fig8}
\end{figure}

To validate the feasibility of jointly simulating tactile images and marker motion images, we leveraged our prior research\cite{SZ24} along with ray tracing to render the depth images obtained by Tacchi. Simultaneously, the method introduced in this work was utilized for simulating marker motion images. We gathered data for pressing, sliding, and rotating actions, covering a comprehensive range of motion types. The results are shown in Figure \ref{fig8}.

The results indicated that Tacchi 2.0 is proficient in jointly simulating various motion forms like pressing, sliding, and rotating across tactile and marker motion images. This also shows that Tacchi 2.0 has great potential for tactile data generation.

%%%%%%%%%%%%%%%%%%%%%%%%%%%%%%%%%%%%%%%%%%%%%%%%%%%%%%%%%%%%%%%%%%%%%%%%%%%%%%%%
\section{Conclusion and Future work}
\label{sec:E}
This paper introduces a marker motion image simulation for pressing, slipping, and rotating based on MPM and camera models. Building upon Tacchi, we propose Tacchi 2.0, a simulator capable of simulating tactile images, marker motion images, and joint images for various motion states including pressing, slipping, and rotating. Besides, since a set of particles is used to represent a marker point, our method has the potential for simulating marker point deformation, thereby obtaining higher information richness than simply considering the displacement of the marker. Experimental results confirm the low computational cost, reliability and effectiveness of this method.

The simulation method based on the MPM and camera model proposed in this paper inherits Tacchi \cite{ZC23} and is adapted to planar optical tactile sensors. With the development of curved optical tactile sensors, this method also has the potential to simulate them. In this paper, we represent planar elastomers with particles, and this method is also applicable to curved elastomers. Considering the focus of dynamic marker motion simulation in this paper, we will verify it in our future work.

In addition, we will continue the upgrade iterations of the Tacchi simulator series to enhance its capabilities. A key focus of the next steps will be integrating the simulation of contact between vision-based tactile sensors and deformable objects\cite{SYH24}, as well as the contact simulation of curved surface vision-based tactile sensors into Tacchi. Concurrently, we will explore integrating Tacchi with mainstream robot simulation platforms like NVIDIA Isaac to produce tactile models for robotic manipulation.

%%%%%%%%%%%%%%%%%%%%%%%%%%%%%%%%%%%%%%%%%%%%%%%%%%%%%%%%%%%%%%%%%%%%%%%%%%%%%%%%
\bibliographystyle{IEEEtran}
\bibliography{IEEEabrv,references}

\end{document}